%% file: root.tex
\documentclass[letterpaper, 10 pt, conference]{ieeeconf}  

\usepackage{graphicx}
\usepackage{amsmath}
\usepackage{epstopdf}
\usepackage[para]{threeparttable}
\usepackage{subfig}
\newcommand{\tabincell}[2]{\begin{tabular}{@{}#1@{}}#2\end{tabular}}  

\IEEEoverridecommandlockouts                              

\title{\LARGE \bf
Gait Graph Optimization: Generate Variable Gaits from One Base Gait for Lower-limb Rehabilitation Exoskeleton Robots
}

\author{Lei Zhang$^1$, Weihai Chen$^{1*}$, Yuan Chai$^2$, Jianhua Wang$^1$, Jianbin Zhang$^1$
\thanks{This work is supported by the National Nature Science Foundation under Grant 61773042, 51675018, 61573047.}
\thanks{$^*$Weihai Chen is the corresponding author.}
\thanks{$^1$Lei Zhang, Weihai Chen, Jianhua Wang, Jianbin Zhang are with the School of Automation science and Electrical Engineering, Beihang University, Beijing, China. e-mail: leizhangbuaa@163.com, whchenbuaa@126.com}
\thanks{$^2$Yuan Chai is with the School of Astronautics, National Key Laboratory of Aerospace Flight Dynamics, Northwestern Polytechnical University,}
\thanks{$^3$https://github.com/leilegelei1/GGO\_CERES.git}
}

\begin{document}

\maketitle
\thispagestyle{empty}
\pagestyle{empty}

\begin{abstract}

The most concentrated application of lower-limb rehabilitation exoskeleton (LLE) robot is that it can help paraplegics "re-walk". However, "walking" in daily life is more than just walking on flat ground with fixed gait. This paper focuses on variable gaits generation for LLEs to adapt complex walking environment. Different from traditional gait generators for biped robots, the generated gaits for LLEs should be comfortable to patients. Inspired by the pose graph optimization algorithm in SLAM, we propose a graph-based gait generation algorithm called gait graph optimization (GGO) to generate variable, functional and comfortable gaits from one base gait collected from healthy individuals to adapt the walking environment. Variants of walking problem, e.g., stride adjustment, obstacle avoidance, and stair ascent and descent, help verify the proposed approach in simulation and experimentation. We open source our implementation$^3$.

\end{abstract}

\section{INTRODUCTION}
Many patients have lost their motor functions in their lower limbs because of diseases like cerebral apoplexy \cite{wuli2003thinking} or accident. Recently, extending LLEs from existing applications including strength augmentation \cite{kazerooni2005exoskeletons} and rehabilitation \cite{murray2014assistive} to assisting locomotion for paraplegics has been a growing interest task of industry. Until now, there has been significant progress in this new field including Ekso, Fourier X1, ReWalk \cite{zeilig2012safety}, HAL \cite{kawamoto2005power} and WAPL \cite{tanabe2013design}. All of these robots can help patients walk smoothly at a flat ground with fixed gait. However, patients have to adjust their gaits for obstacle avoidance \cite{sergey2016study} or stair ascent and descent \cite{jatsun2017footstep}. It is the main issue to develop effective approaches for adapting complex walking environment so that LLEs can be used in daily life.

Considering that paraplegics can hardly provide walking power by themselves, the control strategy of LLEs developed for paraplegics is trajectory tracking control in general. Therefore, the reference joint trajectory for lower-limb, which we call "gait" is quite essential. Traditionally, LLEs simply collect gaits from healthy individuals as the reference joint trajectory, such as Mina \cite{raj2011mina} and Fourier X1. However, this strategy would break down when facing with dimensions obstacles. For different users and different environments, the gait of every walking cycle can be diverse. Unfortunately, It is impossible to record all the gaits from healthy individuals walking in every condition. Hence, an algorithm that can generate functional and variable gaits is quite essential.

\begin{figure}[htbp]
	\setlength{\abovecaptionskip}{0.cm}
	\setlength{\belowcaptionskip}{-0.cm}
	\centering
	\includegraphics[width=\linewidth]{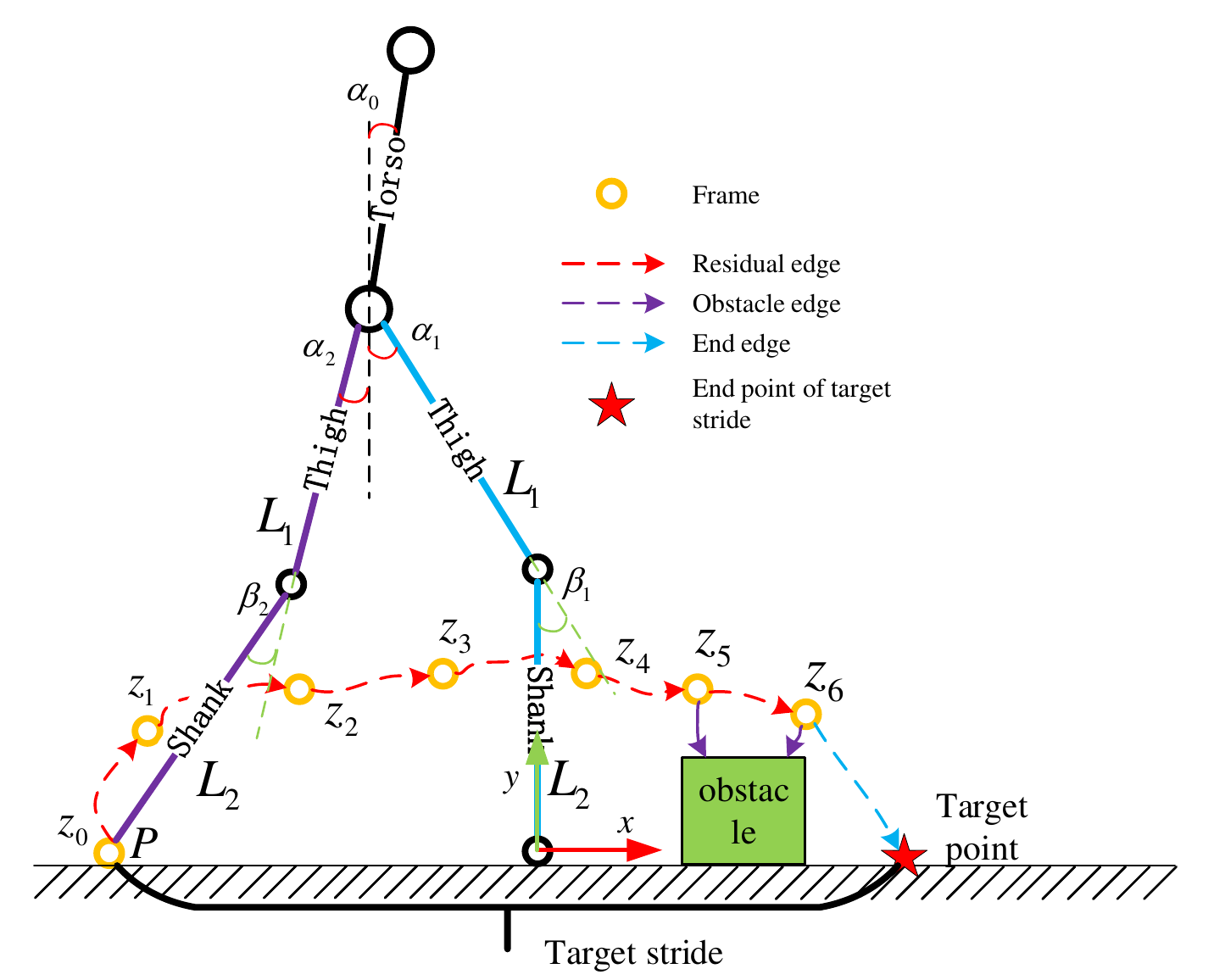}
	\caption{Simplified model of LLE.}
	\label{fig:robot}
\end{figure}
One may face with three kinds of conditions that need to adjust his gait: 1) Adjust his stride to step over a pit. 2) Adjust his gait for obstacle avoidance. 3)Adjust his gait for stair ascent and descent. Although all these three conditions have been researched in robot arms \cite{hoffmann2009biologically}\cite{park2008movement} or biped robots \cite{zhao2014human}\cite{powell2012motion}, these approaches can not be applied in LLEs directly. These methods did not consider whether the gaits are comfortable for paraplegics. Jatsun et al. proposed a strategy to avoid obstacles for LLEs \cite{sergey2016study}. However, this article only considered the trajectory of the endpoint of swinging leg to avoid the obstacle, and the joint angle was solved by the inverse kinematics, which indicates that the comfort of generated gaits can not be guaranteed. Until now, significant progress has been made in the stair ascent and descent strategies for LLEs \cite{ekelem2015preliminary}\cite{taketomi2012stair}. However, both works did not discuss gaits generated online. Sergey et al. proposed an optimization-based algorithm \cite{jatsun2017footstep} for stair ascent, and the joint trajectory was solved by inverse kinematics, which is similar to \cite{sergey2016study}. Recently, Zhong et al. proposed an algorithm that generates smooth gait for stair ascent and descent with LLEs \cite{zhong2019motion}. However, the height of stairs is estimated by the first step, which indicates that this algorithm will not work for stairs with different heights of stair edges. Without the prior knowledge of the stair's height, the height of the first step should be set high enough to step over an uncertain stair. Moreover, without the referenced gait from a healthy person, the comfort of gait can not be guaranteed.

Actually, the basic geometry of obstacles and stairs can be estimated by depth cameras and lidars \cite{cong2008stairway}\cite{zhong2011stairway}\cite{bansal2010lidar}. Now we assume that the geometry of obstacle and stair is obtained, and there exists one gait collected from healthy individuals walking on flat ground, which we call it "base gait". Generally speaking, the base gait is comfortable for patients. If variable gaits can be generated from the base gait through minor modification, the comfort of generated gaits can be guaranteed. Inspired by the pose graph optimization \cite{carlone2015initialization}, we propose gait graph optimization, which can generate gaits from the base gait through minor modification. Besides, we show how to generate gaits with the proposed algorithm for stride adjustment, obstacle avoidance, and stair ascent and descent in detail.

\section{Related Work}
\textbf{Pose graph optimization} \cite{carlone2015initialization} (PGO) is a state-of-the-art formulation for loop closure problem in SLAM. In the SLAM framework, one odometry is adopted to estimate the transformation between two adjacent frames. While there exists minor Gaussian noise in the estimated transformations, the little error on transforms would accumulate to a large error on the final estimated pose.  A loop closure detector (LCD) is utilized to detect the closed-loop relationship of two frames and estimate the transformation between them. In the pose graph, the nodes are estimated poses of frames, the weak edges are the estimated transformations from odometry, and the strong edge is the estimated transformations from LCD. Finally, the Gaussian noise on the weak edges can be eliminated with the guidance of strong edges by solving the graph. The main idea of PGO is to utilize the transformation between closed-loop frames as prior knowledge to eliminate the minor noise between every two adjacent frames. Inspired by the inverse process of PGO, if we add errors to the strong edge deliberately, we can also use PGO to add minor and uniform errors to each weak edge in order to generate a totally different trajectory. Compared with real trajectory, every transformation between two adjacent frames in the generated trajectory is quite similar, but the whole trajectories are totally different. This idea can be transferred to generate variable gaits for LLEs from one base gait.

\section{Kinematics Model \& Base Gait}
\subsection{Kinematic model of LLEs}
The kinematic model of LLEs are simplified as a five-link model, as shown in Fig.\ref{fig:robot}, where the blue leg and purple leg denote the supporting leg and swinging leg. A body coordinate system is adopted to describe our robot model. The endpoint of supporting leg is chosen as the original point, the forward direction as the positive direction of $x$ axis, and the negative direction of gravity as the positive direction of $y$ axis. $L_1$ and $L_2$ denote as the length of thigh and shank. $\alpha$ is the angle between the thigh and vertical direction. $\beta$ is the angle between the shank and thigh. Moreover, $\alpha_1$ and $\beta_1$ are in position direction in Fig.\ref{fig:robot}. The subscript of $_1$ and $_2$ for $\alpha$ and $\beta$ denote the joint angles of supporting leg and swinging leg, respectively. Note that in this paper, we only distinguish the supporting leg and the swinging leg but do not distinguish the left and right leg. By switching the role of swinging leg and supporting leg in turns, the robot can walk normally. The endpoint of swinging leg $P$ can be described as:
\begin{equation}\label{Equ.transform}
\left[                 
\begin{array}{ccc}  
P_x \\
P_y \\ 
\end{array}
\right]                
= T
\left[                 
\begin{array}{ccc}  
L_1 \\
L_2 \\ 
\end{array}
\right]
\end{equation}
where $T$ is:
$$
\left[               
\begin{array}{ccc}  
-sin(\alpha_1)+sin(\alpha_2) & -sin(\alpha_1-\beta_1)+sin(\alpha_2-\beta_2)\\ 
cos(\alpha_1)-cos(\alpha_2) & cos(\alpha_1-\beta_1)-cos(\alpha_2-\beta_2)\\ 
\end{array}
\right] 
$$

\subsection{Base Gait}
\label{2.B}
\begin{figure}[htbp]
	\setlength{\abovecaptionskip}{0.cm}
	\setlength{\belowcaptionskip}{-0.cm}
	\centering
	\includegraphics[width=\linewidth]{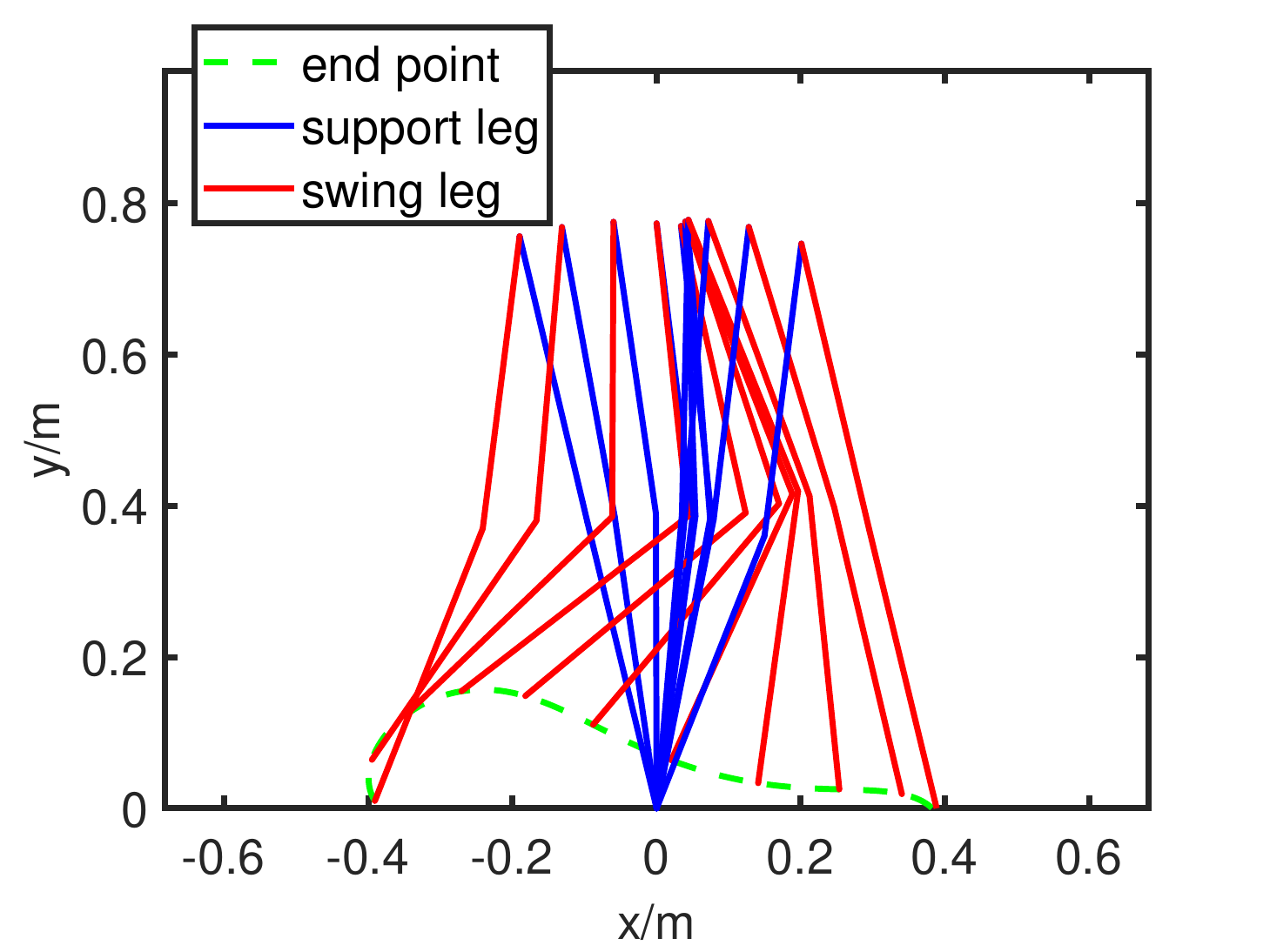}
	\caption{Base gait.}
	\label{fig:base}
\end{figure}

The trajectory waveform of hip and knee joint during gait cycles are periodic and can be collected from healthy individuals. According to the Fourier series, all periodical signal can be decomposed into trigonometric series. Refer to \cite{GuiTowards}, a simply and accuracy gait trajectory is composed of two sine functions
\begin{align}
\left\{                 
\begin{array}{ccc}   
L_{knee}&= 11.35+23.69sin(2\pi t+1.02)\\
&+ 11.35 + 18.54sin(4\pi t+0.41)\\
L_{hip}&= 3.76 + 12.94sin(2\pi t -0.29)\\ 
&+ 3.76 + 4.78sin(4\pi t -0.64)\\
\end{array}
\right.               
\label{equ:trajectory}
\end{align}
Here, we choose this gait as the base gait. By sampling $m$ frames from $0.89s$ to $1.39s$, the trajectory of the swinging leg from the beginning of swinging to the end of a swing can be recorded, as shown by the red line in Fig.\ref{fig:base}. The blue line denotes the trajectory of the supporting leg from the beginning of supporting to the end of supporting by sampling $m$ frames from $1.39s$ to $1.89s$. Here, ``frame" denotes a group of all joint angles of LLEs at sampling time. A frame is described as $\mathbf{z}(n)=[\alpha_1(n),\beta_1(n),\alpha_2(n),\beta_2(n)]^T$, where $n$ denotes the $n_{th}$ sampling in $m$ frames.

\section{Methodology}

\subsection{Algorithm Overview}
As mentioned in \ref{2.B}, we sample $m$ frames as the base gaits. From Equ.\ref{Equ.transform}, we can calculate point $P=[P_x,P_y]^T$ from frame $\mathbf{z}$
\begin{equation}
P=f(\mathbf{z})
\end{equation}
where $f$ denotes the mapping function frame $\mathbf{z}$ to the coordinate of point $P$. Then, define the whole base gait containing $m$ frames as $\mathbf{Z}=\{\mathbf{z}(1),\mathbf{z}(2),...\mathbf{z}(m)\}$ and the adjacent error of $\mathbf{Z}$ as:
\begin{align}
\mathbf{AE}(\mathbf{Z}) &= \{\mathbf{z}(2)-\mathbf{z}(1),\mathbf{z}(3)-\mathbf{z}(2),...,\mathbf{z}(m)-\mathbf{z}(m-1)\}\notag\\
&=\{\mathbf{e}_z(1),\mathbf{e}_z(2),...,\mathbf{e}_z(m-1)\}
\end{align}
where $\mathbf{e}_z(n)$ denotes the error between two adjacent frame $\mathbf{z}(n)$ and $\mathbf{z}(n+1)$ in base gait. The endpoint of frame $i$ is denoted as $P(i)=[P_x(i),P_y(i)]^T$. Then, we can build a graph whose node is the frames $\mathbf{z}$ and the weak edges among nodes are the adjacent errors $\mathbf{e}$, which is quite similar to the pose graph. This kind of edges is called \textbf{``residual edge"}.

Furthermore, we need to generate variable gaits to step over obstacles and adjust the stride through solving the graph. For these proposes, some \textbf{``obstacle edges"} are added into the graph to measure the relative pose between the endpoint of swinging leg and obstacles. Then, we add an \textbf{``end edge"} into the graph, which measures the distance between endpoint $P$ in the last frame of one gait and the target foothold. The overview of the graph is shown in Fig.\ref{fig:robot}

We call our proposed algorithm gait graph optimization(GGO). It is obvious that PGO is trying to eliminate the estimation error between adjacent frames to eliminate the resulting error while GGO is trying to add little error into every frame to generate a totally different gait. GGO is the inverse process of PGO.

\subsection{Residual Edge}
\label{sec:res}
It is assumed that the generate gaits also contain $m$ frames, the $n_{th}$ frame in generated gait is $\textbf{g}(n)$ and the residual error between two adjacent frames is $\textbf{e}_g(n)$. The generated gait is $\textbf{G}=[\textbf{g}(1),\textbf{g}(2),...,\textbf{g}(m)]$. With residual edge, the difference between generate gaits and base gait would be well-proportioned propagated to all frames. Hence, if only comparing one residual error between two adjacent frames $\textbf{e}_g(n)$ and $\textbf{e}_z(n)$ in both generated gaits and base gait, they would be quite similar. Thus the comfort of generated gait can be guaranteed. When walking with both generated gaits and base gait, patients won't feel quite different from frame $n$ to $n+1$. Then, define the residual edges as a minimization problem:
\begin{align}
\label{equ:res_1}
\underset{\textbf{G}}{min}\sum_{i=1}^{m-1}||(\textbf{g}(i+1)-\textbf{g}(i))-(\textbf{z}(i+1)-\textbf{z}(i))||^2
\end{align}

We add a constraint into Equ.\ref{equ:res_1} to ensure $\beta_1$ and $\beta_2$ stay positive for negative angle of the knee may hurt patients. Then, the modified minimization problem is written as:
\begin{equation}
\label{equ:res_2}
\underset{\textbf{G}}{min}\sum_{i=1}^{m-1}\{||\textbf{e}_g(i)-\textbf{e}_z(i)||^2+\lambda \sum_{j=1}^{2}ReLU(-\beta_j)\}
\end{equation}
where $\lambda$ is a gain of the constraint on knee's angle to ensure the second term in Equ.\ref{equ:res_2} can generate enough gradient. $ReLU$ is an activation function that is often used in deep learning:
\begin{align}
ReLU(x)=
\left\{                 
\begin{array}{ccc}   
x,& if\ x>0\\
0,& if\ x<=0
\end{array}
\right.               
\label{equ:no}
\end{align}
With $ReLU$, positive $\beta_1$ and $\beta_2$ won't make any influence to this graph, while negative will generate gradient.

\subsection{Obstacle Edge}
\label{sec:obs}
\begin{figure}[htbp]
	\setlength{\abovecaptionskip}{0.cm}
	\setlength{\belowcaptionskip}{-0.cm}
	\centering
	\includegraphics[width=0.9\linewidth]{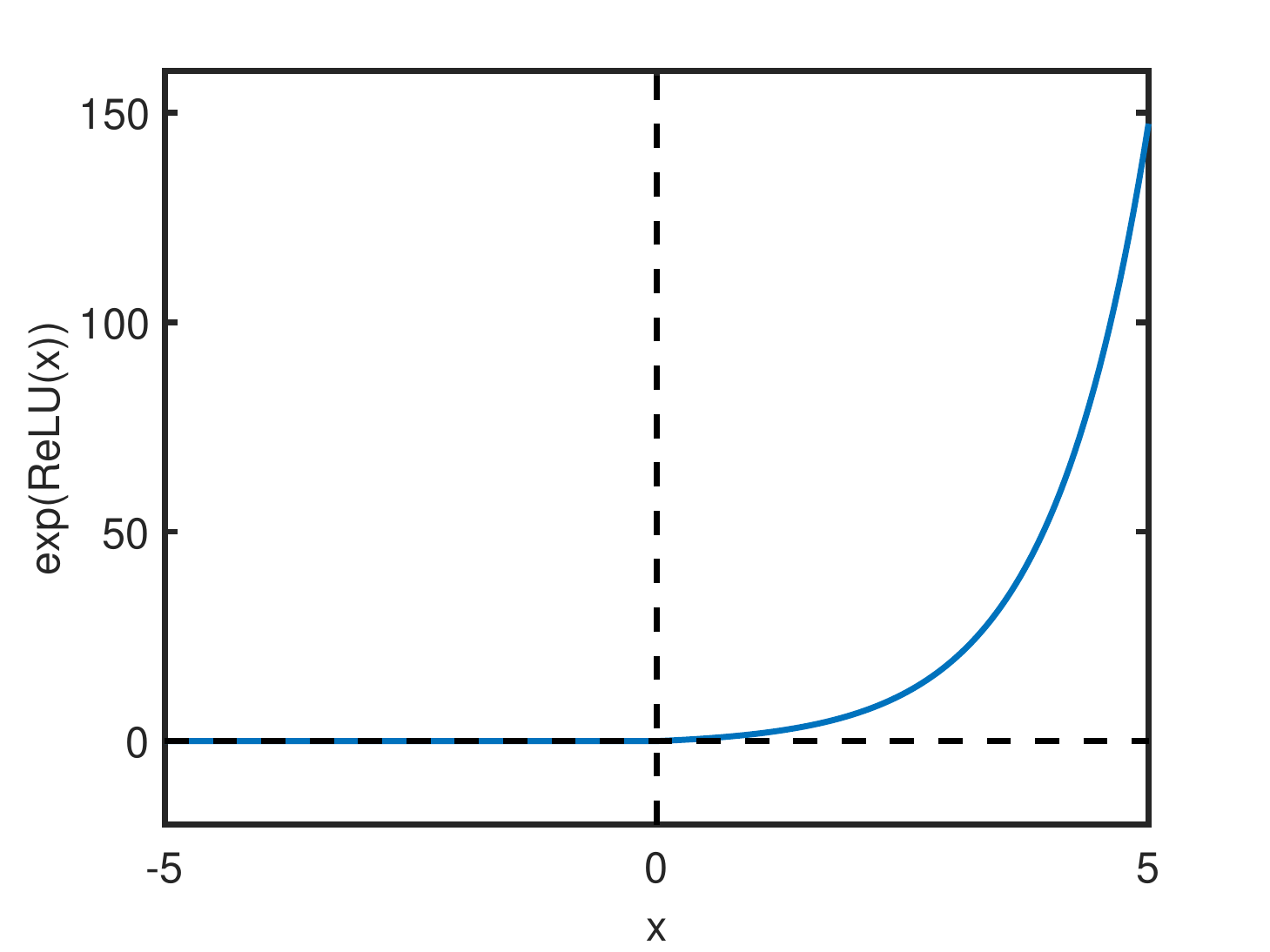}
	\caption{The function $exp(ReLU(x))$.}
	\label{fig:exp}
\end{figure}

The obstacle edge is added into the graph to help LLEs to step over a obstacle object as shown in Fig.\ref{fig:robot}. By adopting the AABB bounding box of obstacles, we can always regard the obstacles as rectangle. It is assumed that an obstacle object is located at the range of $[x_s,x_e]$ and the height of obstacle is $h$. Then, the obstacle avoidance problem can be simplified as:
\begin{align}
\label{equ:obss_1}
P_y(i)>h, \forall i, P_x(i)\in [x_s,x_e]
\end{align}
Assumed $\mathbf{O}=\{\textbf{g}(i),\textbf{g}(i+1),...,\textbf{g}(j)\}$ as a gather of all frames in a generate gait whose $P_x$ is in $[x_s,x_e]$, the obstacle edge can be written as:
\begin{align}
\label{equ:obs_1}
\underset{\textbf{O}}{min}\sum_{k=i}^{j}||exp(\gamma \cdot ReLU(h+\delta-P_y(k))))-1||^2
\end{align}
where $\delta$ is a hyper-parameters to strengthen the robustness of our algorithm. Note that we only optimize the frames in $\mathbf{O}$. $ReLU(h+\delta-P_y(k))$ indicates that only when $P_y(k)<h+\delta$, the frame $\textbf{g}(k)$ will be optimized. $\gamma$ is a gain to enlarge the error. Besides, we apply an $exp$ activation function on the error. As shown in Fig.\ref{fig:exp}, when $x$ grows, $exp(ReLU(x))$ grows rapidly. As the frames that take part in obstacle edge are much fewer than that in residual edge, such that the active function can ensure the obstacle edge provide enough gradient to ensure the gait can step over the obstacle.

\subsection{End Edge}
\label{sec:end}
The end edge which only influences the last frame of generated gaits is adopt to set a proper foothold of the swinging leg. It can be used for stride adjustment or stair ascent and descent. Now we set a target foothold for the swinging leg as $F=[F_x,F_y]^T$. The end edge can be denoted as
\begin{align}
\label{equ:end_1}
min\{\omega||f(\mathbf{g}(m))-F||^2\}
\end{align}
where $\omega$ is the gain of end edge and g(m) is the last frame of the gait. The angle between ground and shank of LLEs should also be considered when swinging leg reaches the foothold. For example, when a healthy person ascends one stair, his swinging leg is approximately perpendicular to the ground in most cases. This found drives us to add a term to constrain the angle $\epsilon$ between ground and shank of swinging leg in last frame. The $\epsilon$ can be calculated by:
\begin{align}
\label{equ:end_2}
\epsilon = \alpha_2(m)-\beta_2(m)
\end{align}
where $\alpha_2(m)$ and $\beta_2(m)$ are the value of $\alpha$ and $\beta$ in the last frame of generated gait. Denoting the target angle as $\epsilon_t$, then the whole end edge is written as:
\begin{align}
\label{equ:end_3}
min\{\omega(||f(\mathbf{g}(m))-F||^2+ ||\epsilon- \epsilon_t||)\}
\end{align}

\subsection{Solving The Graph}
Adding all the edges mentioned in \ref{sec:res}, \ref{sec:obs} and \ref{sec:end}, we can build a full graph for GGO algorithm. Then, solving the gait graph is equal to solve a minimization problem:
\begin{align}
\underset{\textbf{G}}{min}
\left\{                 
\begin{array}{ccc}   
&\sum_{i\in \mathbf{G}}\{||\textbf{e}_g(i)-\textbf{e}_z(i)||^2\\
&+\lambda \sum_{j=1}^{2}ReLU(-\beta_j)\}+\\
&\sum_{k\in \mathbf{O}}\{||exp(\gamma \cdot ReLU\\
&(h+\delta-f(\textbf{g}(i)))))-1||^2\}+\\
&\{\omega(||f(\mathbf{g}(m))-F||^2+ ||\epsilon- \epsilon_t||)\}
\end{array}
\right\}               
\label{equ:min}
\end{align}
Based on different applications, edges can be added or removed from the graph. By using the non-linear optimization algorithm like L-M or DogLeg, the graph can be solved. 

\section{Experiment}
In this section, we will show various applications of GGO on the LLEs in both simulations and experiments. These applications contain three parts. 1) Generation of gaits with different strides. 2) Generation of gaits for dimensions obstacles avoidance. 3) Generation of gaits for stair ascent and descent. All these gaits are generated form one base gait introduced in Fig.\ref{fig:base}. All graphs are solved by Ceres Solver\cite{agarwal2012ceres}, and the non-linear optimization algorithm is chosen as DogLeg. And $L_1=L_2=0.39(m)$.

\subsection{Generate Gaits with Different Strides}\label{diff_stride}
\begin{figure}[htbp]
	\setlength{\abovecaptionskip}{0.cm}
	\setlength{\belowcaptionskip}{-0.cm}
	\centering
	\includegraphics[width=0.8\linewidth]{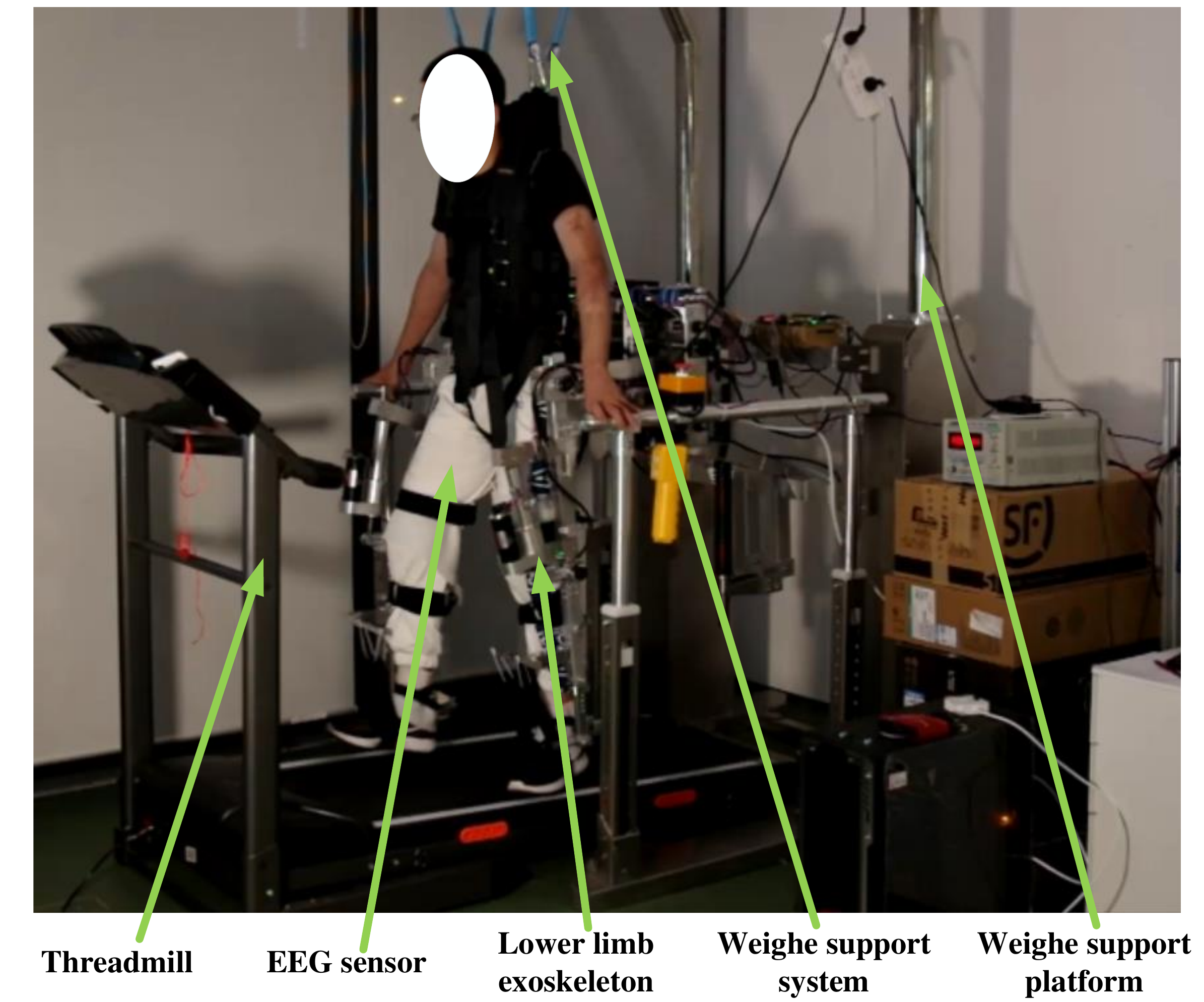}
	\caption{Experimental platform.}
	\label{fig:platform}
\end{figure}
Define the stride as the distance between point $P$ in the first frame and point $P$ the last frame in one gait. The stride of base gait is $0.78(m)$. Now we show how to generate a gait with stride of $0.975(m)$. Here, only "residual edge" and "end edge" should be added into the graph and we won't add the constraint of the angle $\epsilon$. The hyper-parameters is set to $\lambda=5, \omega=5$. Then, we can generate a gait with stride of $0.975(m)$ in two steps:
\begin{itemize}
	\item \textbf{First}, initialize $\mathbf{G}$ with $\mathbf{Z}$, and set the target foothold of swinging leg to $F=[0.4875, 0]^T(m)$. Then, build the graph and solve the graph by DogLeg algorithm. 
	\item \textbf{Second}, define the gait generate by first step as $\mathbf{G^1}$ and the last frame of $\mathbf{G^1}$ as $\mathbf{g^1(m)}$. Then, initialize all frames in $\mathbf{G}$ as  $\mathbf{g_m^1}$ and also set $F=[0.4875, 0]^T(m)$. Finally,  build the graph and solve the graph by DogLeg algorithm. 
\end{itemize}
\begin{figure}[htbp]
	\setlength{\abovecaptionskip}{0.cm}
	\setlength{\belowcaptionskip}{-0.cm}
	\centering
	\includegraphics[width=\linewidth]{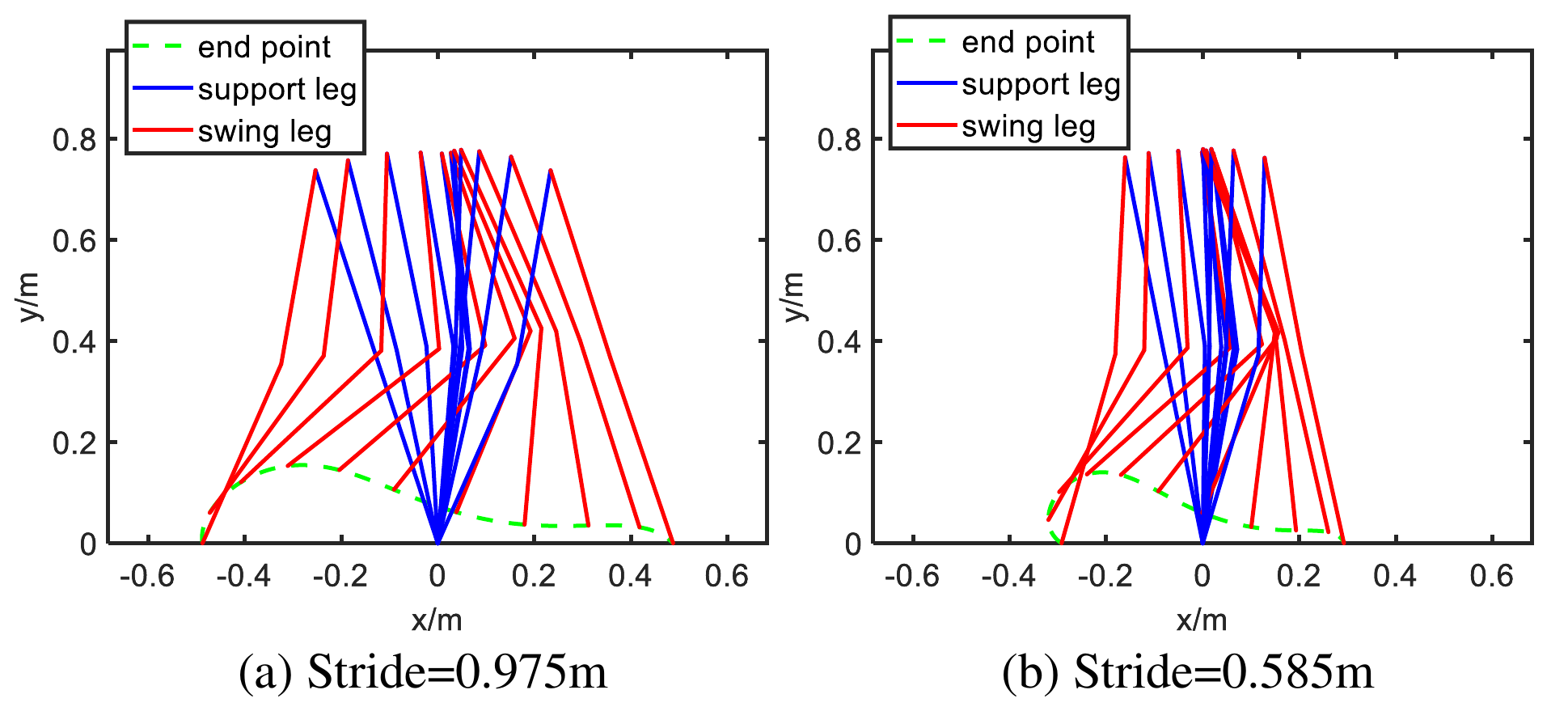}
	\caption{Generated gait with different stride.}
	\label{fig:diff_stride}
\end{figure}

With the same process, we can generate a gait with a stride of $0.585(m)$. The gaits with stride of $0.975(m)$ and $0.585(m)$ are shown in Fig.\ref{fig:diff_stride}(a) and Fig.\ref{fig:diff_stride}(b). Note that there exists a role switching of support leg and swinging leg in the two steps.

\begin{figure*}[htbp]
	\setlength{\abovecaptionskip}{0.cm}
	\setlength{\belowcaptionskip}{-0.cm}
	\centering
	\includegraphics[width=\linewidth]{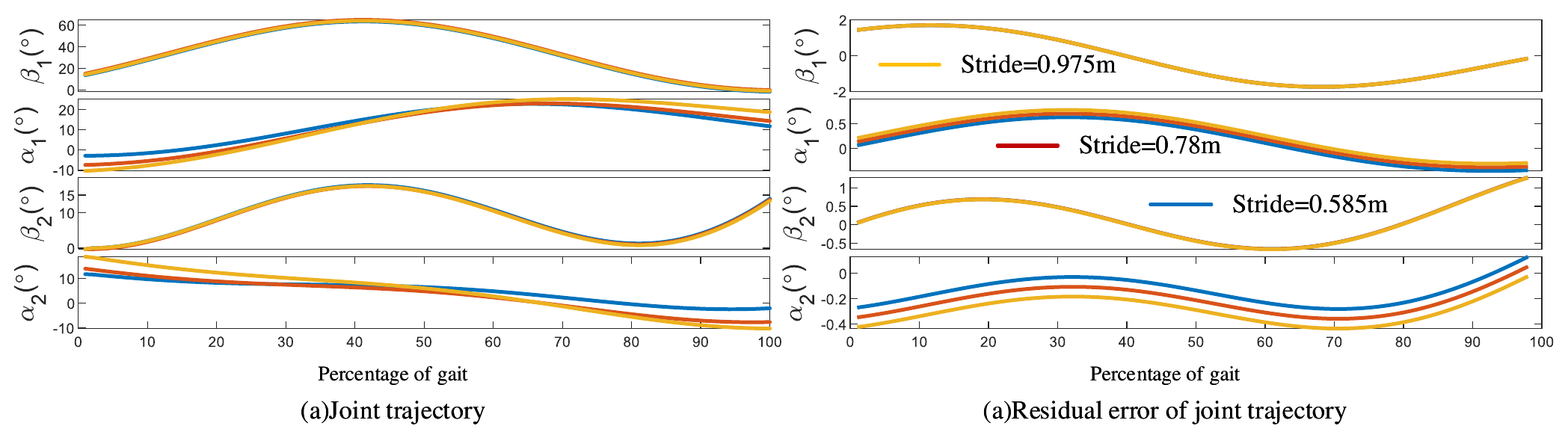}
	\caption{Generated gait for stir descents.}
	\label{fig:stridetrajectory}
\end{figure*}

To verify the effectiveness of the proposed algorithm for the real LLEs, we designed an experiment based on the experimental platform shown in Fig.\ref{fig:platform}. The lower limb rehabilitation robotic system consists of a treadmill, a weight support platform, and a lower limb exoskeleton. By detecting the activity degree of muscle in thigh using an EEG sensor, the gait stride of $0.975(m)$,$0.78(m)$ and $0.585(m)$ would switch autonomously. The higher activity degree corresponds to longer stride. The trajectory of each joint in generated gait is shown in Fig.\ref{fig:stridetrajectory}(a) and the residual error of every frame is shown in Fig.\ref{fig:stridetrajectory}(b).
From Fig.\ref{fig:stridetrajectory}(a), the generated gait is smooth enough to be tracked and quite similar to base gait. From Fig.\ref{fig:stridetrajectory}(b), it is found that the huge different strides in the whole gaits are caused by the accumulation of the little difference in every frame. As the differences of residual error among the three gaits in each frame are significantly small, the feeling of patient moving from $i_{th}$ frame to $i+1_{th}$ is quite similar. Thus the generated gait is also comfortable for patients when doing rehabilitation training. To analysis the difference between generated gaits and base gait, we calculate the average value of absolute joint angle error between generated gait and base gait, as shown in Tab.\ref{tab:abs_err}. The biggest difference between the two generated gaits and base gait are the knee angle of the swinging leg. However, the largest value is also smaller than $3^\circ$, such that the comfort performance of generated gaits can be guaranteed. 
\begin{table}[htbp]
	\setlength{\abovecaptionskip}{0.cm}
	\setlength{\belowcaptionskip}{-0.cm}
	\caption{The average absolute error between generated gait and base gait}
	\centering
	\scriptsize
	\setlength\tabcolsep{6pt} 
	\label{tab:abs_err}
	\begin{threeparttable}[t]
		\begin{tabular}{| l || *{6}{ c |}  }
			\hline
			\textbf{Joint} & ${\beta_1}^\circ$ & ${\alpha_1}^\circ$ & ${\beta_2}^\circ$ & ${\alpha_2}^\circ$\\
			\hline \hline
			Stride=$0.585(m)$ & 1.4508 & 1.9439 & 0.2686 & 2.3270
			\\
			\hline
			Stride=$0.975(m)$ & 0.7991 &  1.9386 &   0.1575   & 2.1009 \\
			\hline
		\end{tabular}
	\end{threeparttable}
\end{table}

\subsection{Generate Gaits for Obstacle Avoidance}
\begin{figure}[htbp]
	\setlength{\abovecaptionskip}{0.cm}
	\setlength{\belowcaptionskip}{-0.cm}
	\centering
	\includegraphics[width=\linewidth]{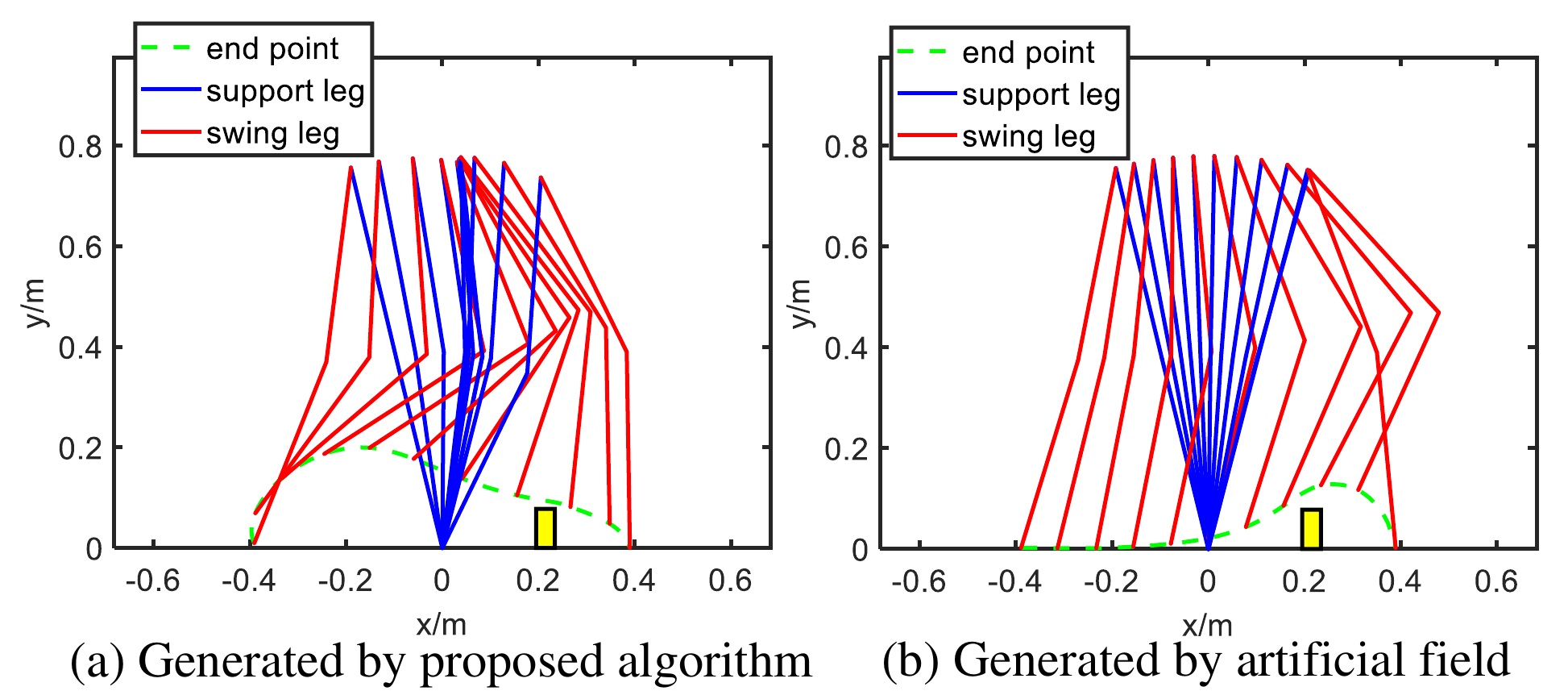}
	\caption{The gait with different strides.}
	\label{fig:crossobstacle}
\end{figure}

First, we provide an example to generate a gait to step over a rectangle obstacle start from $[0.2,0]^T(m)$ to $[0.25,0.8]^T(m)$ as shown in Fig.\ref{fig:crossobstacle}. We add all the three types of edges into the graph. Specifically, we do not add the constraint of the angle $\epsilon$. The hyper-parameters of the graph is set to $\lambda=5, \omega=5m, \gamma=4, \delta=0.02$, and the target foothold of swinging leg is $F=[0.39, 0]^T(m)$. The gather $\mathbf{O}$ contains all frame whose $P_x$ in the range of $[0.18,0.22](m)$. By using DogLeg to solve the graph, the generated gait is shown in Fig.\ref{fig:crossobstacle}(a).

As a comparison, we generate another gait using the artificial filed algorithm \cite{lee2003artificial}. The trajectory of endpoint $P$ is generated using the artificial filed algorithm, and the joint angles calculate the joint angle in all frames by solving the inverse kinematics. The joint trajectory and residual of joint trajectory are shown in Fig.\ref{fig:obstacletrajectory}(a) and Fig.\ref{fig:obstacletrajectory}(b). Fig.\ref{fig:obstacletrajectory} shows that the gait generated by GGO much more smooth than the gait generated by AF. The great performance of GGO is because of the introduction of the residual edge. With the residual edge, the base gait plays as a teacher to guide the generation of new gait, such that the generated gait is similar to base gait. Moreover, we try to conduct an ablation study to demonstrate the influence of residual edge. However, the experimental results show that GGO can not generate proper gait at all without the residual edge in the graph.

In order to compare these two algorithms further, the time cost and the comfort performance are chosen as the evaluation indicator. Comfort performance is measured based on ZMP. Zero moment point (ZMP) \cite{kajita2003biped} is an important indicator for the balance of legged robot. When ZMP is in the support area of a robot, the robot can keep balance itself. When ZMP is further from the foothold, the robot gets more unbalanced. Although patients in the LLEs can keep balance with two walking sticks which are used to enlarge the supporting area so that the ZMP can be included in the supporting area, a far distance between ZMP and foothold may cause patients uncomfortable as patients would exert more force on the walking stick in order to keep balance. The ZMP can be expressed by the following equation:
\begin{align}
x_{zmp} = \frac{\sum_{i=1}^{n}m_i[(\frac{d^2y_i}{dt^2}+g)x_i-y_i\frac{d^2x_i}{dt^2}]}{\sum_{i=1}^{n}m_i(\frac{d^2y_i}{dt^2}+g)}
\end{align}
where $x_{zmp}$ is the distance between ZMP and support point. $n$ is the link number of robot, in our problem $n=4$. $[x_i,y_i]^T$ is the center point of the $i_{th}$ link, and $g$ is the gravity constant. $m_i$ is the mass of $i_{th}$ link. In our experimental configuration, both the length of thigh and shank as $0.39(m)$ and both the two links weight $7(kg)$. Moreover, the torso link is not considered. The obstacles in the two tables are shown in Tab.\ref{tab:obstacles}. The average $x_{zmp}$ of different gaits are shown in Tab.\ref{tab:zmp} and the time cost to solve the gait graph proposed in this paper is shown in Tab.\ref{tab:time}.
\begin{table}[htbp]
	\setlength{\abovecaptionskip}{0.cm}
	\setlength{\belowcaptionskip}{-0.cm}
	\caption{The experimental configuration.}
	\centering
	\scriptsize
	\setlength\tabcolsep{6pt} 
	\label{tab:obstacles}
	\begin{threeparttable}[t]
		\begin{tabular}{| l | *{6}{ c |} }
			\hline
			Obstacles & $x_s(m)$ & $x_e(m)$ & $h(m)$ \\
			\hline \hline
			obstacle 1 & 0.2 & 0.25 & 0.08 \\
			\hline
			obstacle 2 &  0.2 &  0.25 &   0.15 \\
			\hline
			obstacle 3 &  0.12 &  0.16 &   0.08\\
			\hline
			obstacle 4 & 0.12 &  0.16 &   0.15 \\
			\hline
			obstacle 5 & -0.33 &  -0.29 &   0.175 \\
			\hline
		\end{tabular}
	\end{threeparttable}
\end{table}

In Tab.\ref{tab:zmp} ``Base gait" denotes the performance of base gait. ``proposed" denotes the performance of our proposed algorithm.``AF" denotes the performance of the Artificial field. In Tab.\ref{tab:time},``iteration" denotes the iteration cycle to solve the graph and``time cost" denotes the time cost of the proposed algorithm. The average $x_{zmp}$ of gait generate by the artificial field algorithm is significantly higher than the gaits generated by the proposed algorithm. In addition, from Tab.\ref{tab:time}, we found that the time cost of solving gaits for different obstacles are similar to each other, and the average time cost is $77.9ms$.
\begin{figure*}[htbp]
	\setlength{\abovecaptionskip}{0.cm}
	\setlength{\belowcaptionskip}{-0.cm}
	\centering
	\includegraphics[width=\linewidth]{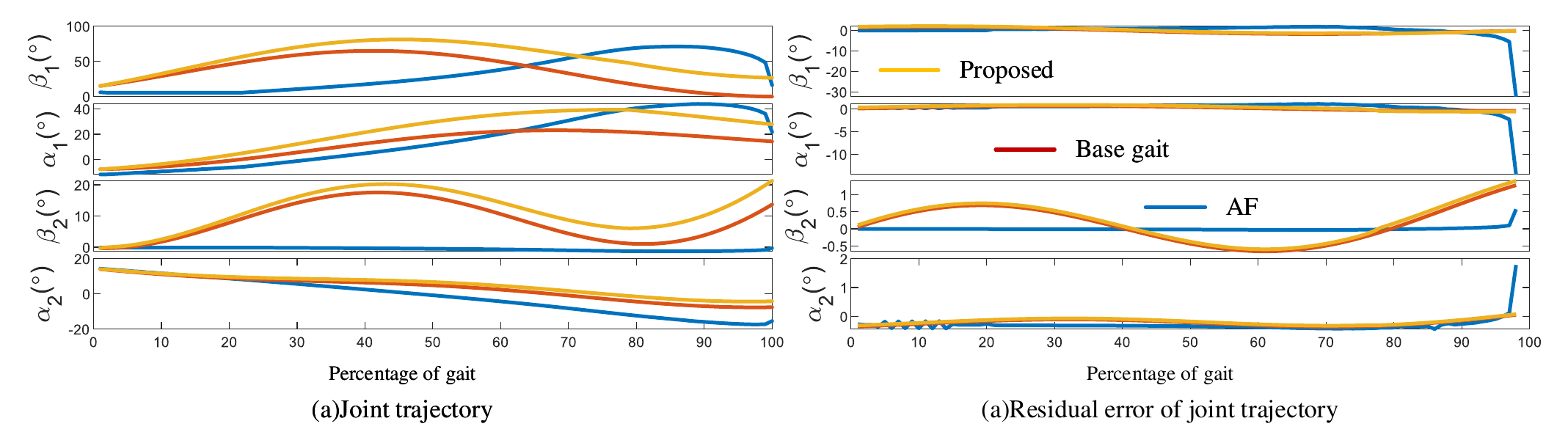}
	\caption{Generated gait for stir descents.}
	\label{fig:obstacletrajectory}
\end{figure*}
\begin{figure*}[htbp]
	\setlength{\abovecaptionskip}{0.cm}
	\setlength{\belowcaptionskip}{-0.cm}
	\centering
	\includegraphics[width=0.9\linewidth]{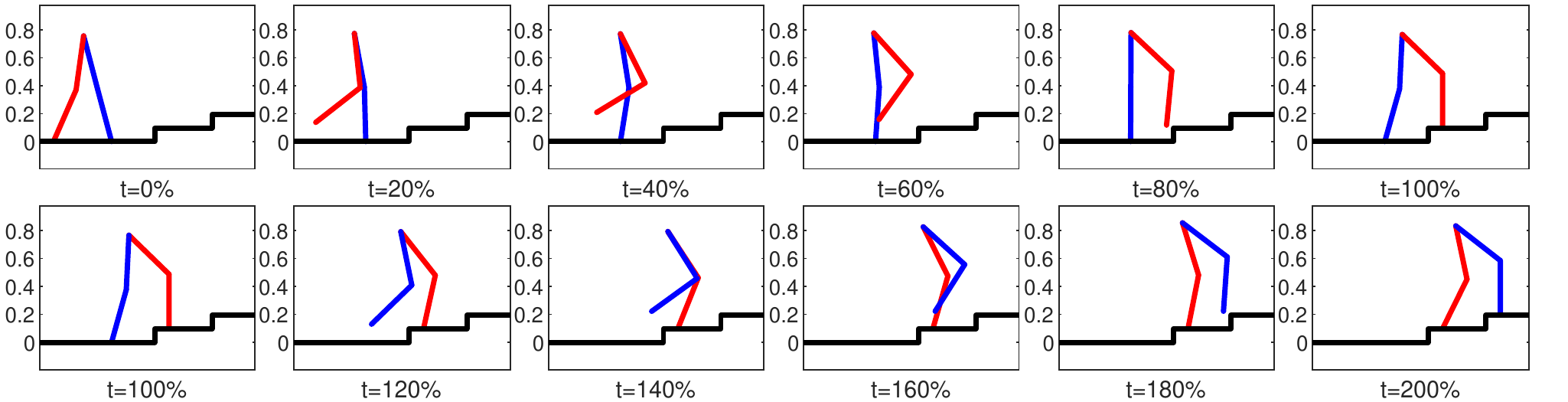}
	\caption{Generated gait for stair ascents.}
	\label{fig:upstair}
\end{figure*}
\begin{figure*}[htbp]
	\setlength{\abovecaptionskip}{0.cm}
	\setlength{\belowcaptionskip}{-0.cm}
	\centering
	\includegraphics[width=0.9\linewidth]{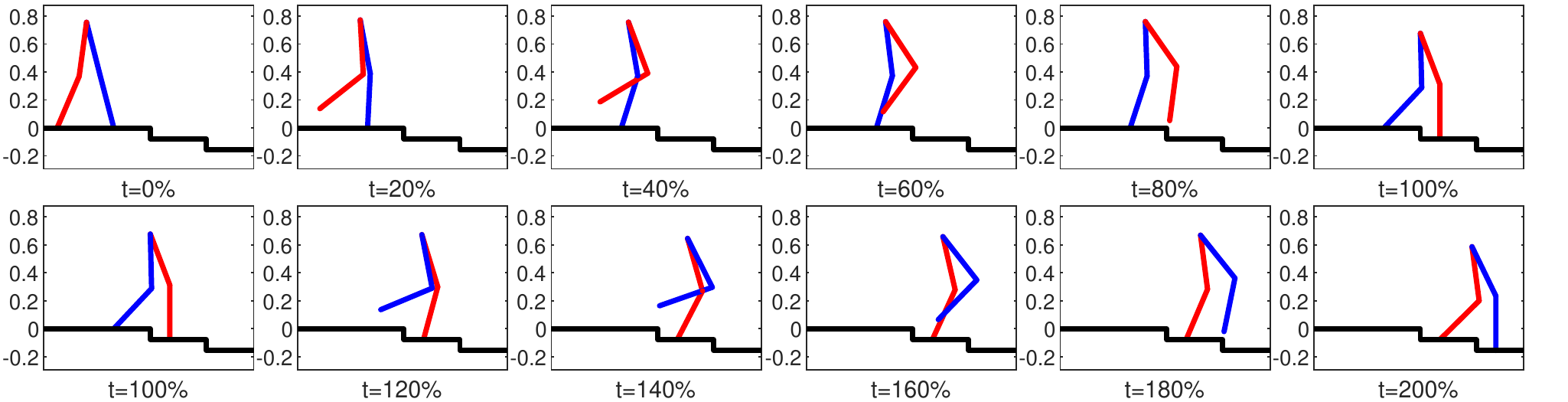}
	\caption{Generated gait for stir descents.}
	\label{fig:downstair}
\end{figure*}

\begin{table}[htbp]
	\setlength{\abovecaptionskip}{0.cm}
	\setlength{\belowcaptionskip}{-0.cm}
	\caption{Average $x_{zmp}$ of different gaits.}
	\centering
	\scriptsize
	\setlength\tabcolsep{6pt} 
	\label{tab:zmp}
	\begin{threeparttable}[t]
		\begin{tabular}{| l | *{6}{ c |} }
			\hline
			\tabincell{c}{Average $x_{zmp}$(m)} & Base gait & Proposed & AF \\
			\hline \hline
			obstacle 1 & 0.0828 & 0.0986 & 0.2188 \\
			\hline
			obstacle 2 &  0.0828 &  0.1204 &   0.2382 \\
			\hline
			obstacle 3 &  0.0828 &  0.0914 &   0.1752\\
			\hline
			obstacle 4 & 0.0828 &  0.1460 &   0.1969 \\
			\hline
			obstacle 5 & 0.0828 &  0.0823 &   0.2515 \\
			\hline
			\textbf{Average} & 0.0828 &  0.0828 & 0.2162\\
			\hline
		\end{tabular}
	\end{threeparttable}
\end{table}

\begin{table}[htbp]
	\setlength{\abovecaptionskip}{0.cm}
	\setlength{\belowcaptionskip}{-0.cm}
	\caption{Time cost of proposed algorithm}
	\centering
	\scriptsize
	\setlength\tabcolsep{6pt} 
	\label{tab:time}
	\begin{threeparttable}[t]
		\begin{tabular}{| l | *{4}{ c |} }
			\hline
			\textbf{} & Iterations & Time Cost(ms) \\
			\hline \hline
			obstacle 1 & 47 & 77.6 \\
			\hline
			obstacle 2 &  47 &  77.3 \\
			\hline
			obstacle 3 &  44 &  76.8 \\
			\hline
			obstacle 4 & 50 &  80.9  \\
			\hline
			obstacle 5 & 35 &  76.9 \\
			\hline
			\textbf{Average} & 44.6 & 77.9 \\
			\hline
		\end{tabular}
	\end{threeparttable}
\end{table}

\begin{figure}[htbp]
	\setlength{\abovecaptionskip}{0.cm}
	\setlength{\belowcaptionskip}{-0.cm}
	\centering
	\includegraphics[width=\linewidth]{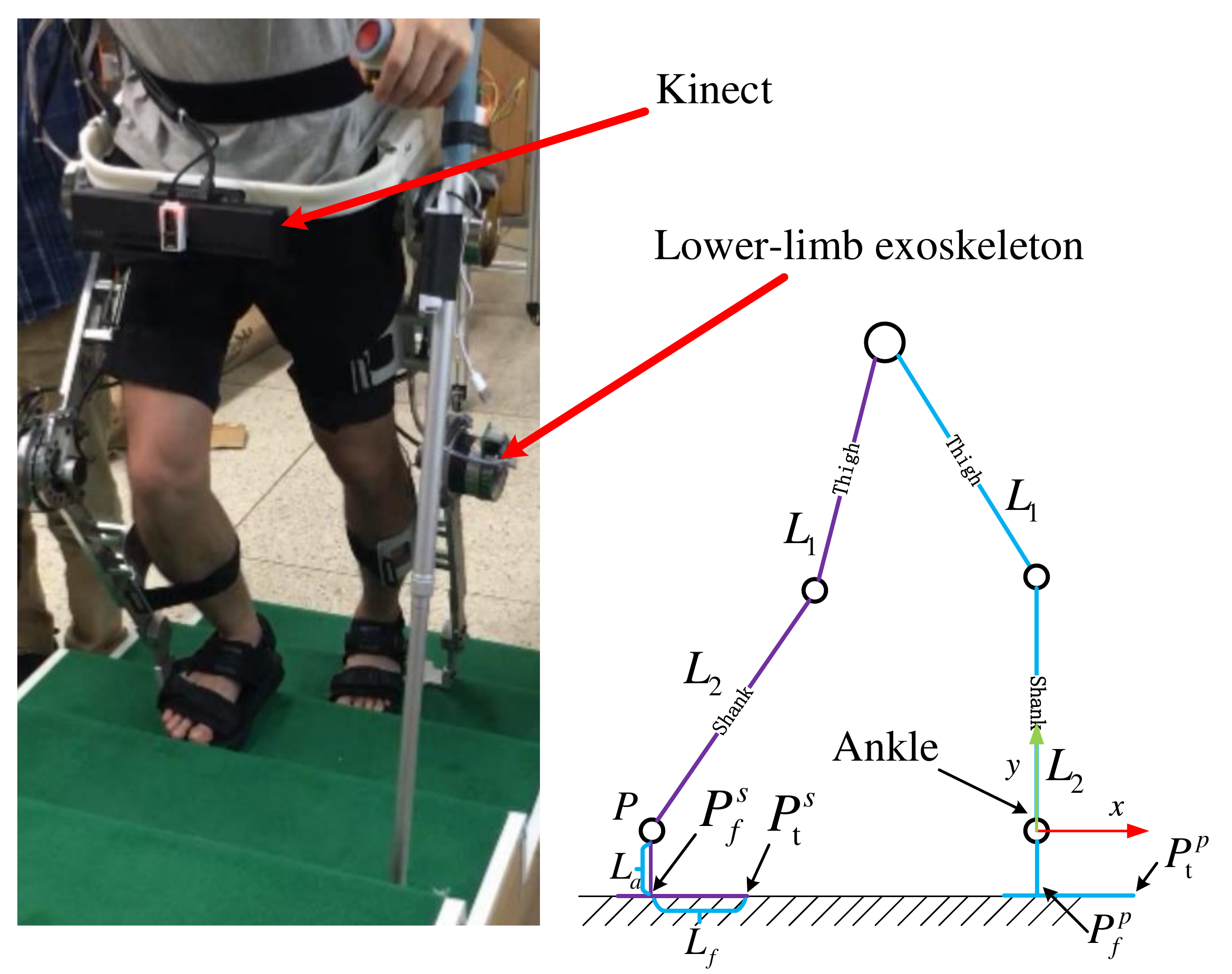}
	\caption{Experimental robot for stair ascent.}
	\label{fig:uprobot}
\end{figure}
\begin{figure}[t]
	\setlength{\abovecaptionskip}{0.cm}
	\setlength{\belowcaptionskip}{-0.cm}
	\centering
	\includegraphics[width=\linewidth]{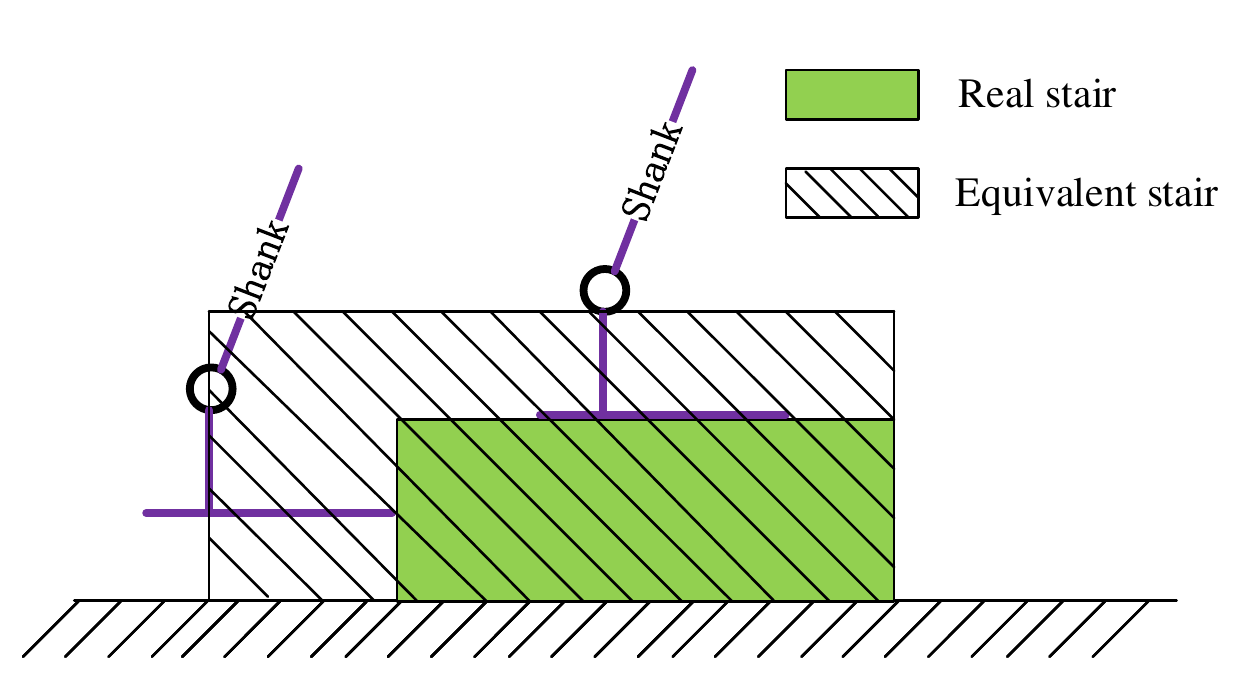}
	\caption{Equivalent stair for point $P$.}
	\label{fig:equivalent}
\end{figure}
\begin{figure}[t]
	\setlength{\abovecaptionskip}{0.cm}
	\setlength{\belowcaptionskip}{-0.cm}
	\centering
	\includegraphics[width=\linewidth]{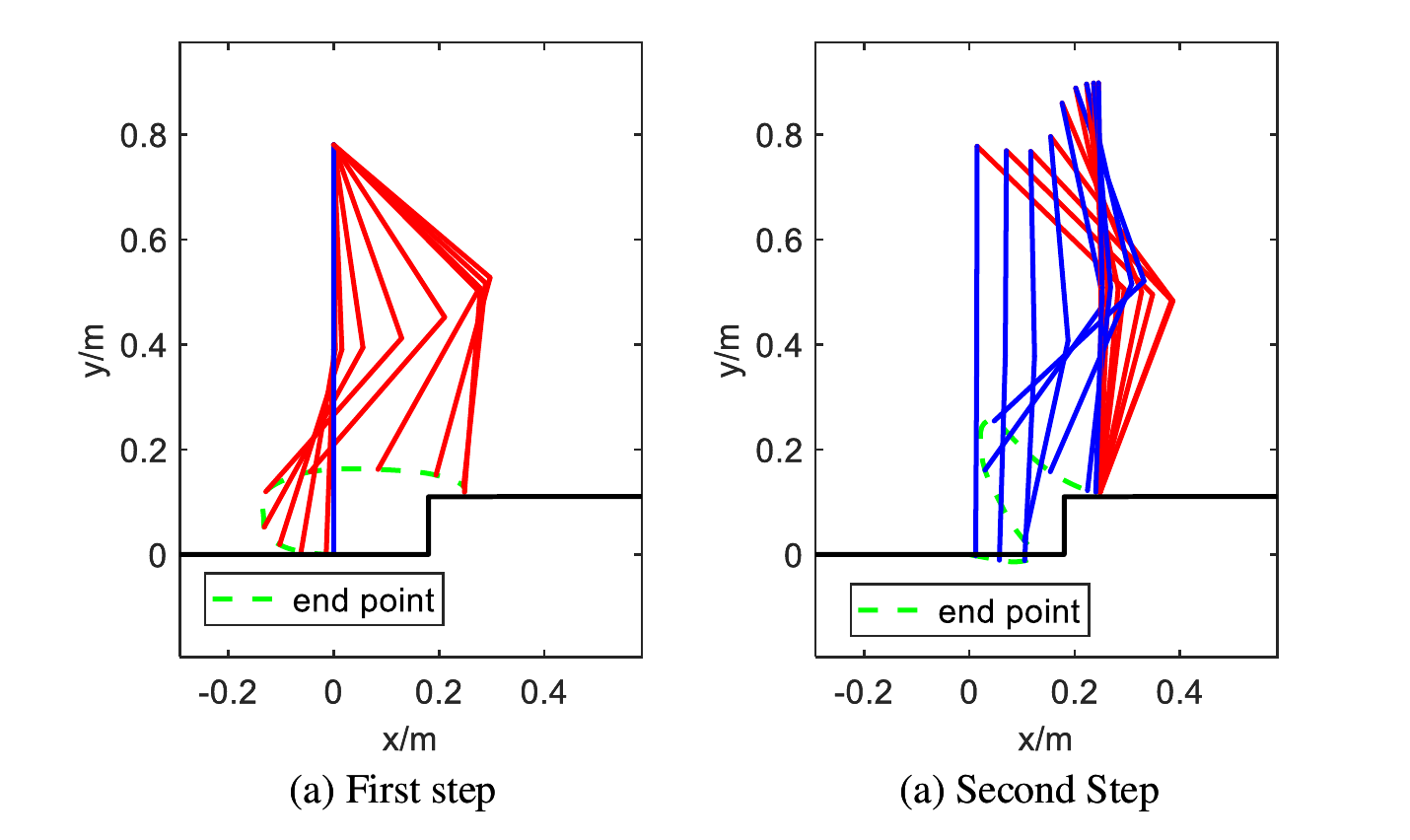}
	\caption{Collected gait for stair ascent.}
	\label{fig:coll}
\end{figure}
\subsection{Generate Gaits for Stair Absent and Descent}
We firstly provide an example to generate a gait for stair ascent. In this example, the stair has two stages, and the rising edges are at $x=0.29(m)$ and $x=0.68(m)$. The height of both stairs is $0.1(m)$. The generated gait is shown in Fig.\ref{fig:upstair}, $t=n\%$ denotes the frame is sampling from the $n\%$ time of the whole gait. To generate the gait shown in Fig.\ref{fig:upstair}, we add all three kinds of edges into the graph. The hyper-parameters are set to $\lambda=5, \omega=5\, \gamma=4, \delta=0.02$ and the target angle $\epsilon_t=90^\circ$. Then, we can generate the gait within two steps:
\begin{itemize}
	\item \textbf{First}, initialize $\mathbf{G}$ with $\mathbf{Z}$, and set the target foothold of swinging leg to $F=[0.39, 0.1]^T(m)$. Then, set the gather $\mathbf{O}$ contains all frame whose $P_x$ is in the range of $[0.28,0.30](m)$, and set the height of obstacle to $h=0.1(m)$. Finally, build the graph and solve the graph by the DogLeg algorithm. Here we can generate the gait to step over the first stage, which is shown in the first row of Fig.\ref{fig:upstair}.
	\item \textbf{Second}, define the gait generate by first step as $\mathbf{G^1}$ and the last frame of $\mathbf{G^1}$ as $\mathbf{g_m^1}$. Then, initialize all frames in $\mathbf{G}$ as  $\mathbf{g_m^1}$. Note that the original point is always set to the endpoint of the supporting leg. The supporting leg and swinging leg has changed after the first step. Then, set the rest of the configuration the same as the first step. We can generate the second step to step over the second stage, which is shown in the second row of Fig.\ref{fig:upstair}.
\end{itemize}

In a similar way, a gait for stair descent can be generated by changing the foothold to$F=[0.39, -0.1]^T(m)$. the results are shown in Fig.\ref{fig:downstair}. As mentioned in \ref{diff_stride}, there exists a role switch of support leg and swinging leg in the two steps.

To verify the effectiveness of the algorithm in the real LLEs, we designed an experiment based on experimental platform shown in Fig.\ref{fig:uprobot}. We adopt a Microsoft Kinect2 to detect the stair model. Here, the real LLEs are a bit little different from the model in Fig.\ref{fig:robot}. The endpoint of swinging leg shown in Fig.\ref{fig:robot} is actually the ankle joint in real robot shown in Fig.\ref{fig:uprobot}. It is assumed that the feet are always parallel with the ground. In Fig.\ref{fig:uprobot}, $P_f^s$ denotes the point of intersection between vertical line from point $P$ to feet of swinging leg, $P_t^s$ denotes the tiptoe of swinging leg. We aim to generate a trajectory for the point $P$ to make sure LLEs can step over the stair and no collisions occur among $P_f^s$, $P_t^s$ and stairs. Hence, the stair with height of $h=h_{height}$ in real-world which starts from $x=x_{stair}$ is equivalent to a stair with height of $h=h_{height}+L_a$ which starts from $x=x_{stair}-L_f$ and for point $P$ as shown in Fig.\ref{fig:equivalent}. Note that the stair model detected by Kinect is also transformed to the body coordinate system whose original point is the ankle point of supporting leg. For safety consideration, we collect a gait trajectory from a healthy person for stair ascent as shown in Fig.\ref{fig:coll}, rather than using the base gait from Fig.\ref{fig:base}. In this gait, one cycle of stair ascent contains two steps, the first step start from $\alpha_1=\alpha_2=\beta_1=\beta_2=0$ as shown in Fig.\ref{fig:coll}(a) and the second step end with $\alpha_1=\alpha_2=\beta_1=\beta_2=0$ as shown in Fig.\ref{fig:coll}(b).

\begin{figure}[t]
	\setlength{\abovecaptionskip}{0.cm}
	\setlength{\belowcaptionskip}{-0.cm}
	\centering
	\includegraphics[width=\linewidth]{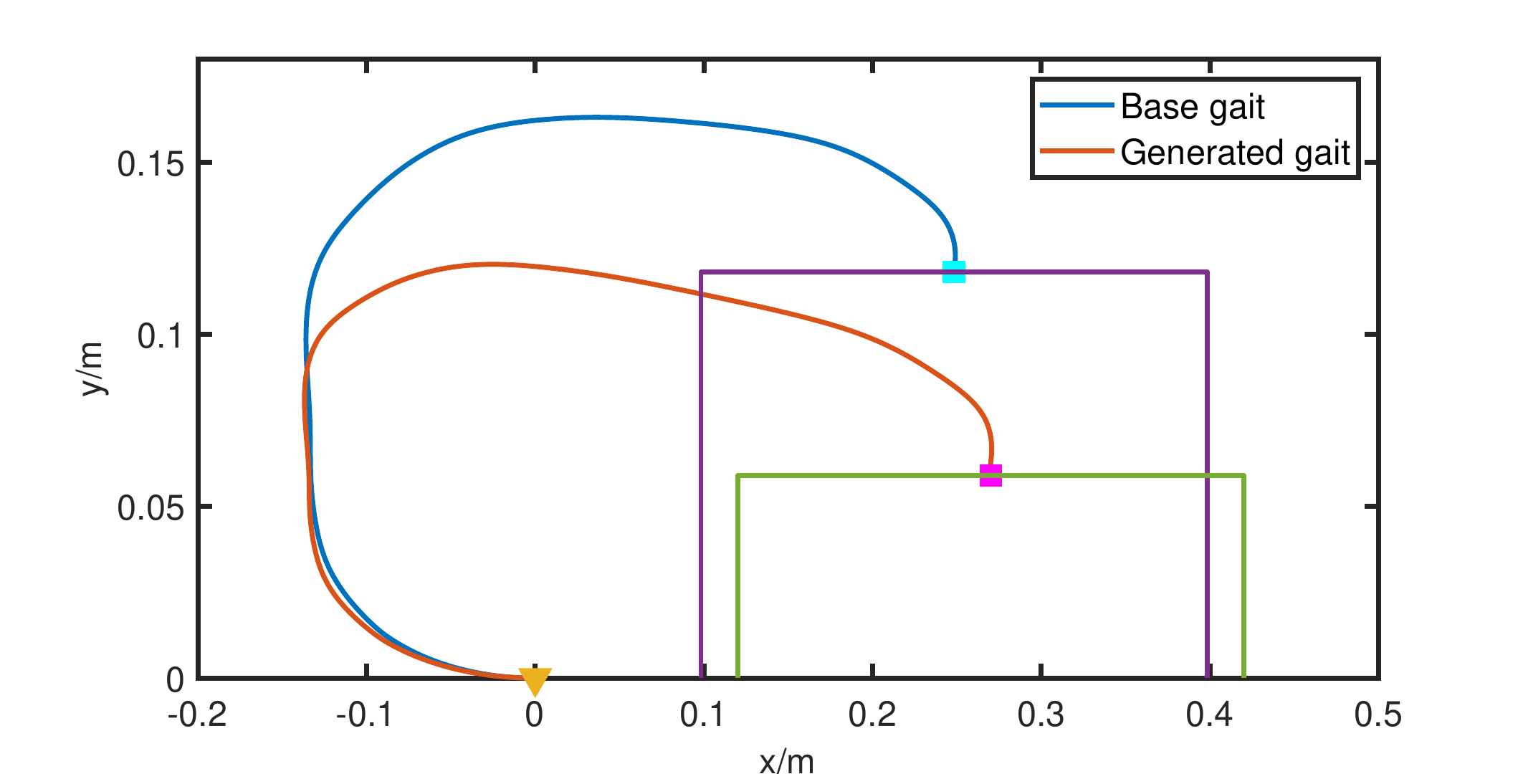}
	\caption{Equivalent stair for point $P$.}
	\label{fig:upgait}
\end{figure}

\begin{figure}[t]
	\setlength{\abovecaptionskip}{0.cm}
	\setlength{\belowcaptionskip}{-0.cm}
	\centering
	\includegraphics[width=\linewidth]{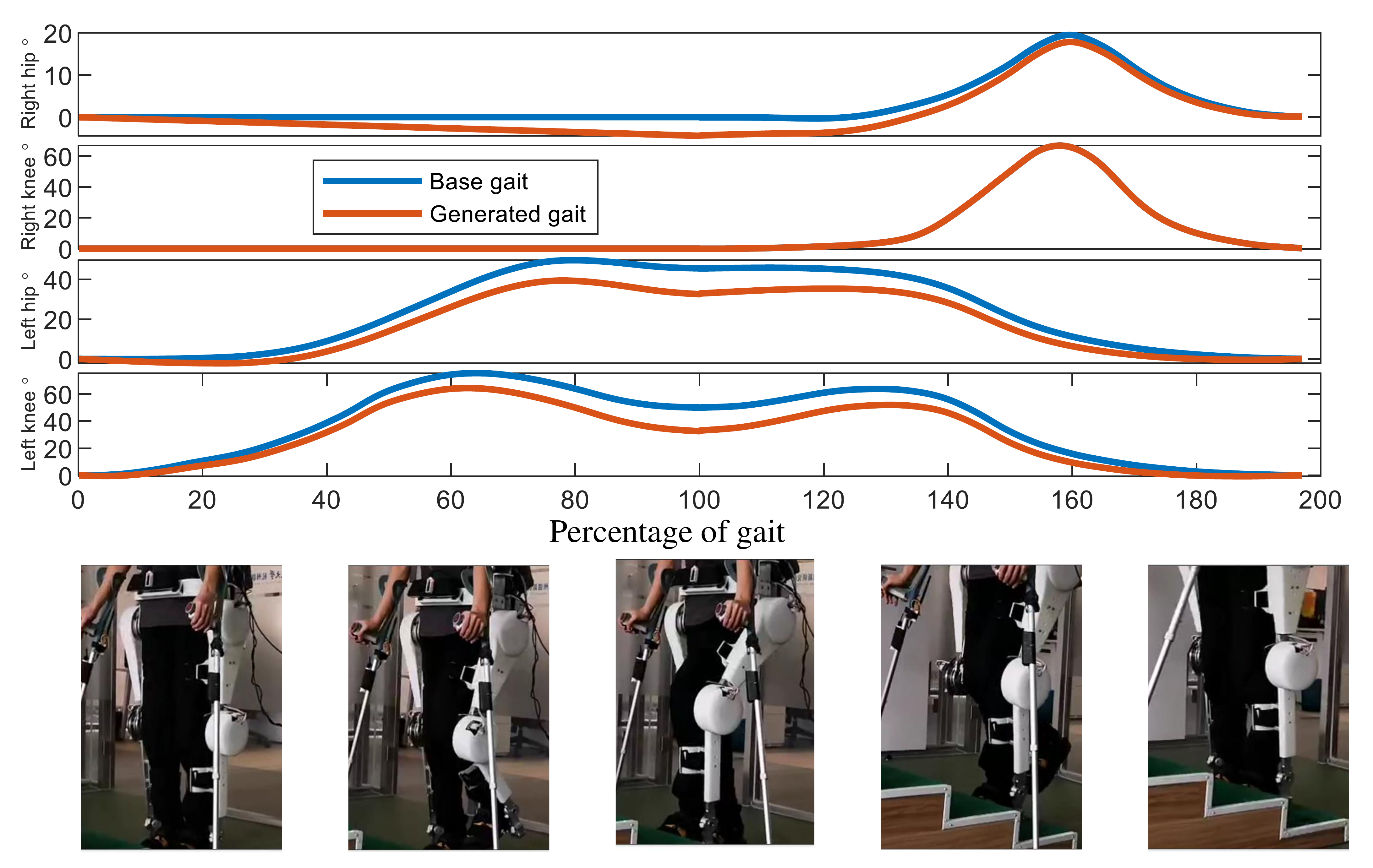}
	\caption{Equivalent stair for point $P$.}
	\label{fig:upstair_trajectory}
\end{figure}

The height of the stair for gait trajectory collection is $0.12$m, and the foothold for the point $P$ is $[0.25, 0.12]^T(m)$. In the experimental configuration, the height of the stair is $0.06m$, so we set the foothold of point $p$ to $[0.27,0.06]^T(m)$. The experimental results are shown in Fig.\ref{fig:upgait} and Fig.\ref{fig:upstair_trajectory}. As shown in Fig.\ref{fig:upgait}, the two curves are the trajectory of point $P$ of base gait and generated gait. The joint trajectory of both legs is shown in Fig.\ref{fig:upstair_trajectory}, the four subgraphs are the angle of the hip joint of the right leg, the angle of knee joint of the right leg, the angle of hit joint of the left leg and the angle of knee joint of the left leg. This result demonstrates that our approach can be extended to generate gait for stair ascent and descent. Moreover, the generated gaits are functional and comfortable. 
\section{Conclusion \& Future Work}
In this paper, we propose a graph-based algorithm to generate variable gaits from one base gait for stride adjustment, obstacle avoidance and stair ascent and descent. The GGO shows the generalization performance for different conditions and its usability by conducting a large number of simulations and experiments. Moreover, the superior performance are demonstrated by comparing against other algorithms used for mobile robots. We open source the C++ implementation to benefit the community. 

Although the proposed algorithm works well in variants applications, there are still many directions to perfect our research in the future. The collision problem including collision volume should be considered in a more precise way, which is simplified in this paper. Besides, the strategy for choosing the foothold of the swinging leg should also be carefully designed.

\bibliographystyle{IEEEtran}
\input{root.bbl}


\end{document}